\documentclass{article}

\usepackage{arxiv}

\usepackage[utf8]{inputenc} 
\usepackage[T1]{fontenc} 
\usepackage{hyperref} 
\usepackage{url} 
\usepackage{booktabs} 
\usepackage{amsfonts} 
\usepackage{nicefrac} 
\usepackage{microtype} 
\usepackage{lipsum}
\usepackage{graphicx}
\usepackage{subfig}
\usepackage{float}
\usepackage{color}
\usepackage{lineno}
\graphicspath{ {./images/} }
\usepackage{tablefootnote}

\usepackage {todonotes} 

\usepackage{tabularx,ragged2e}
\usepackage{appendix}
\usepackage{booktabs} 
\usepackage{graphicx} 
\usepackage{array}
\usepackage{subcaption}
\usepackage[affil-it]{authblk} 
\usepackage{refcount} 
\usepackage{caption}

\usepackage{titlesec}
\titleformat{\section}{\normalfont\Large\bfseries}{}{0em}{}
\title{}

\author[1]{Minghua Wu}
\author[2]{Javier Conde}
\author[2]{Pedro Reviriego}
\author[1]{Marc Brysbaert}

\affil[1]{Department of Experimental Psychology, Ghent University}
\affil[2]{Information Processing and Telecommunications Center (IPTC), Universidad Politécnica de Madrid, Spain}

\begin{document}

\title{How does fine-tuning improve sensorimotor representations in large language models?}
\maketitle



\begin{abstract}
Large Language Models (LLMs) exhibit a significant "embodiment gap," where their text-based representations fail to align with human sensorimotor experiences. This study systematically investigates whether and how task-specific fine-tuning can bridge this gap. Utilizing Representational Similarity Analysis (RSA) and dimension-specific correlation metrics, we demonstrate that supervised fine-tuning with human ratings drives a targeted reorganization of the model’s semantic space rather than a simple global improvement of existing patterns. We observed a near-zero correlation between the performance rankings of base and fine-tuned models, indicating that the corrective recalibration was not uniform across concepts. Furthermore, the results show that while sensorimotor improvements generalize robustly across languages and related sensory-motor dimensions, they are highly sensitive to the learning objective, failing to transfer across two disparate task formats. This research underscores the remarkable plasticity of LLMs, suggesting that their internal representations can be steered toward more embodied, grounded patterns through targeted supervision.
\end{abstract}



\section{Introduction}
In recent years, large language models (LLMs) have made remarkable breakthroughs in natural language processing and have demonstrated impressive capabilities in understanding and generating language. These advances have prompted researchers to use LLMs to generate ratings of word characteristics. Comparisons with human ratings have shown high correlations for variables such as word concreteness, valence, and arousal \cite {martinez_ai-generated_2025}, familiarity \cite{brysbaert_moving_2024}, and age of acquisition \cite{green_crowdsourced_2025, sendin_combining_2025}. These results indicate that estimates generated by LLMs for these variables are a useful addition to complement human ratings.

Since LLMs primarily rely on language training, \cite{conde_psycholinguistic_2025} and \cite{xu_large_2025} investigated whether the models could provide accurate estimates of embodied experiences despite lacking such experiences. As expected, LLMs achieved human-like representation levels in non-sensorimotor features; however, their performance in sensorimotor features, including sensory experiences and action-related concepts, significantly lagged behind human capabilities. This representational gap underscores a fundamental limitation of current language models—the absence of direct interaction with the physical world results in systematic deficiencies in grasping concepts related to sensory perception and motor actions.

Further investigation by \cite{xu_large_2025} comparing GPT-3.5 and GPT-4 demonstrated that models that receive multimodal input (text and images) during training exhibited more human-like representation patterns when processing visually relevant dimensions. These results imply that models require rich multimodal information during training, similar to the way humans gain experience through bodily interactions with the environment, to potentially achieve human-level representation in sensorimotor domains.

\cite{sendin_combining_2025} and \cite{green_crowdsourced_2025} presented an alternative method to improve LLM performance. They showed that age-of-acquisition ratings improved substantially when a model was fine-tuned using 2,000 human ratings. During this process, the model first provides an estimate, is given the human rating, and is asked to update the weights so that the model's outcome aligns with the human rating. Since training multimodal large models poses significant challenges due to the need for substantial amounts of multimodal data and substantial computational resources, fine-tuning could be a more effective method for adapting models and enhancing their specific capabilities.

In this study\footnote{Data and code for this paper are available at: \url{https://github.com/Minghua5/Sensorimotor-Representation-in-LLMs.git}}, we explore whether fine-tuning offers a viable pathway to address the fundamental issue of the disconnect between language models and the physical world, specifically focusing on sensorimotor representation. Using GPT-4o-mini as our base model and creating several fine-tuned variants, we conduct a systematic evaluation. Our primary criterion for evaluating the quality of a model's sensorimotor representation is its similarity to human-derived representations. By comparing the similarity between model and human representations before and after fine-tuning, we assess the efficacy of the method: if the fine-tuned model demonstrates greater representational alignment with humans, we conclude that fine-tuning is effective; otherwise, we deem it ineffective.

We conduct a systematic, multi-level analysis of this similarity. First, at the overall structural level, we employ Representational Similarity Analysis (RSA) to construct semantic spaces defined by various sensorimotor features and examine the relative positioning of different concepts within these spaces. Second, we disaggregate the analysis into specific dimensions, investigating how the model-human similarity varies across individual sensorimotor features. This allows us to identify on which specific dimensions fine-tuning is effective or ineffective. Third, we drill down to the level of individual concepts to determine whether improvements from fine-tuning are consistent across all concepts or limited to a subset, thereby pinpointing potential areas of persistent failure. Furthermore, by varying the type of training data (e.g., different task formulations) and the language of the data, we investigate the generalizability and scalability of the fine-tuning approach across tasks and languages.

This investigation deepens our understanding of model plasticity and its alignment with human cognition. It also provides a theoretical foundation for developing more balanced and comprehensive artificial intelligence semantic systems. Through a series of rigorously controlled experiments, we aim to systematically evaluate the efficacy, generalizability, and limitations of fine-tuning methods for enhancing the sensorimotor representation of LLMs.

\section{Results}

\subsection{Overall Structural Alignment: Representational Similarity Analysis (RSA)}

\begin{figure}[htbp]
    \centering
    \subfloat[Tested on EN set]{
        \includegraphics[width=0.45\textwidth]{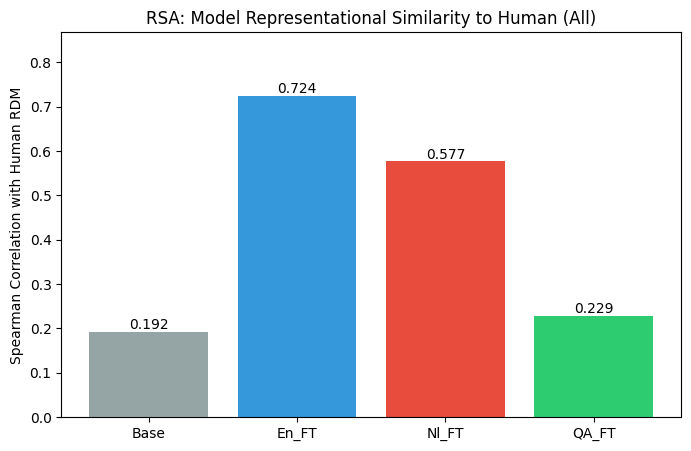}
        \label{fig:rsa_all} 
    }
    \hfill 
    \subfloat[Tested on NL set]{
        \includegraphics[width=0.45\textwidth]{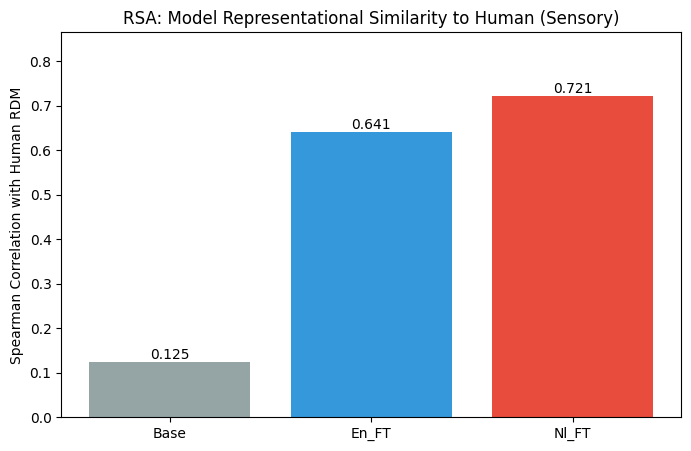}
        \label{fig:dutch_rsa}
    }
     \hfill 
    \subfloat[]{
        \includegraphics[width=0.45\textwidth]{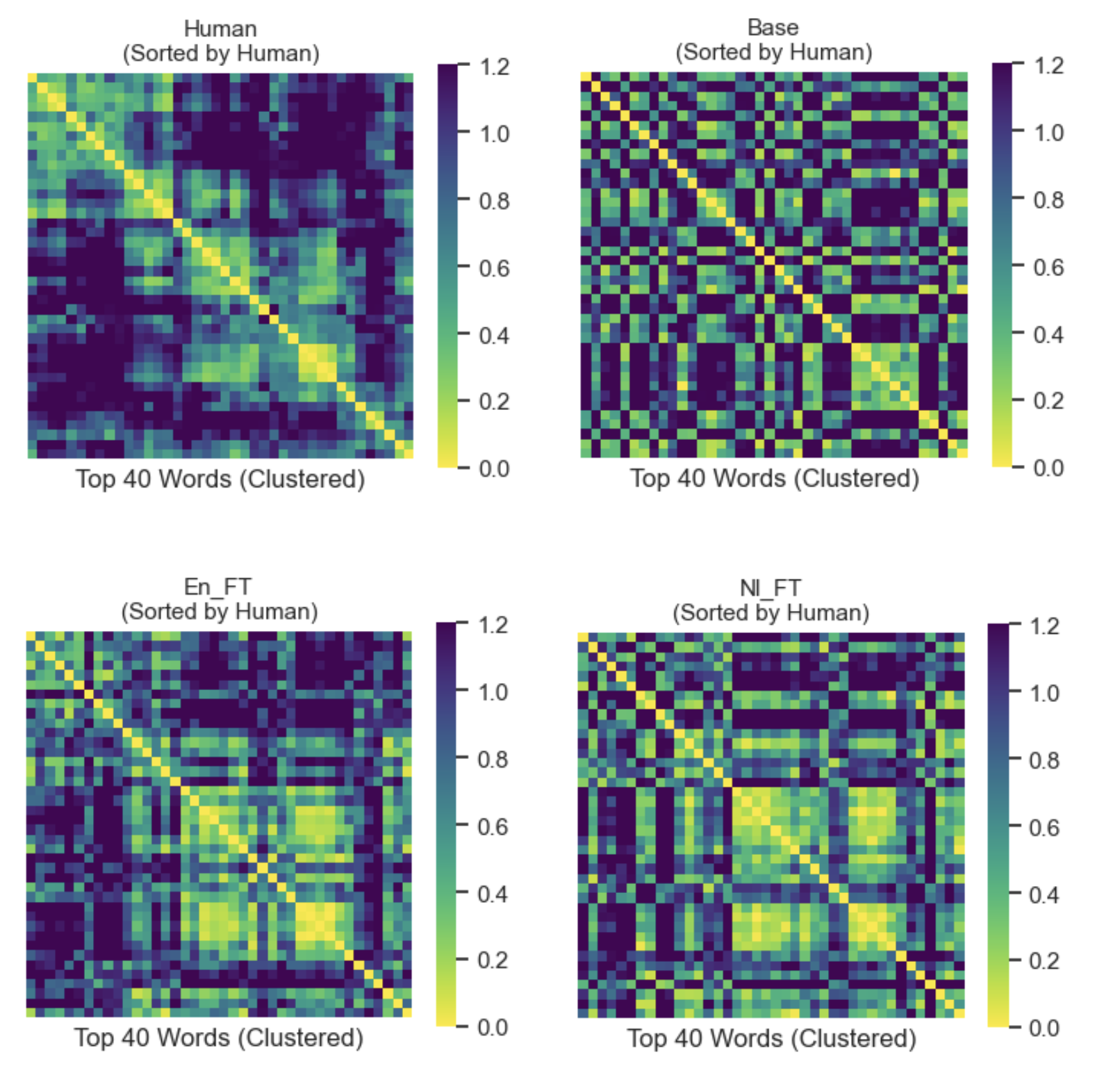}
        \label{fig:rdm}
    }
    \hfill 
    \subfloat[]{
        \includegraphics[width=0.45\textwidth]{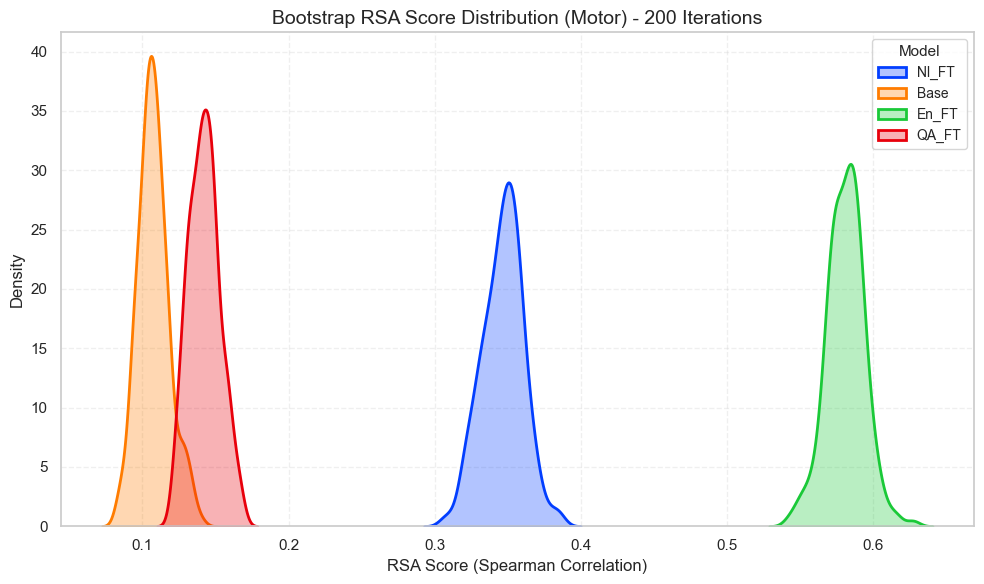}
        \label{fig:test_motor}
    }
    \caption{Structural Alignment of Sensorimotor Representations Before and After Fine-tuning. Bar chart showing the overall Spearman’s $\rho$ correlation between model-derived and human-derived representational similarity matrices (RDMs) for all sensorimotor dimensions on the English evaluation set (a) and for sensory dimensions on the Dutch evaluation set (b). (c) Representational Distance Matrices (RDMs) for human ratings and model representations, based on a representative subset of 40 English words (the full dataset contains 1572 concepts). (d) Density distributions of Spearman’s $\rho$ for motor dimensions on the English set, derived from 200 bootstrap resampling iterations.}
    \label{fig:rsa_results} %
\end{figure}

To assess the fundamental impact of fine-tuning on the overall structure of the models' sensorimotor representations, we first computed the similarity (Spearman rank correlation) between the representational similarity matrices (RDMs) of the base model (GPT-4o-mini), the fine-tuned models, and the human RDM. The results (Fig.\ref{fig:rsa_results}) consistently demonstrate that fine-tuning significantly enhances the overall structural alignment between LLM representations and human representations of sensorimotor experience.

As shown in Fig.\ref{fig:rsa_all}, all three fine-tuned models (En\_FT, Nl\_FT, QA\_FT) achieved significantly higher RSA similarity to human ratings on sensorimotor representations compared to the base model. More intuitive evidence comes from the direct comparison of RDMs (Fig.\ref{fig:rdm}). The human RDM exhibits a clear block-like structure, whereas the base model's RDM appears more diffuse. After fine-tuning, the RDMs of En\_FT and Nl\_FT models became more structured. To quantify the statistical reliability of this improvement, we performed a bootstrap resampling test with 200 iterations. The results confirmed that the differences between each fine-tuned model and the base model were statistically significant. This verifies that the improvements conferred by fine-tuning are robust and not due to random variation.

Critically, the improvements demonstrate clear generalizability across both task formulations and languages. Fine-tuning on Dutch ratings (Nl\_FT) substantially improved performance on English concepts ($\rho$ increased from 0.192 to 0.577), and conversely, fine-tuning on English ratings (En\_FT) led to strong improvement on Dutch concepts ($\rho$ increased from 0.125 to 0.641, see Fig\ref{fig:dutch_rsa}). This bidirectional boost confirms that the acquired structural knowledge transfers across linguistic boundaries. However, this transfer is not complete. Optimal performance on a target language was consistently achieved by the model fine-tuned on that same language (Dutch test: Nl\_FT $\rho$ = 0.721 vs. En\_FT $\rho$ = 0.641; English test: En\_FT $\rho$ = 0.724 vs. Nl\_FT $\rho$ = 0.577). This establishes a hierarchy of efficacy: fine-tuning with matched language data is most effective, followed by cross-lingual transfer, with cross-task transfer (QA\_FT) yielding the smallest, though still significant, gains on the rating prediction task.

An important finding concerns models fine-tuned on incomplete data. Although the Nl\_FT and QA\_FT models were trained exclusively on sensory dimension ratings, their representations of motor features also showed significant improvement during testing. Specifically, the RSA similarity for motor features increased from 0.105 (base) to 0.140 for QA\_FT and to 0.345 for Nl\_FT, with bootstrap tests confirming the significance of these gains (Fig.\ref{fig:test_motor}). This suggests that fine-tuning on sensory attributes can facilitate a partial, indirect learning of related motor representations, indicating a degree of interconnected generalization within the sensorimotor semantic space.

\subsection{Dimension-Specific Improvements and Variances}
\begin{figure}[htbp]
    \centering
    \subfloat[Tested on EN set]{
        \includegraphics[width=0.8\textwidth]{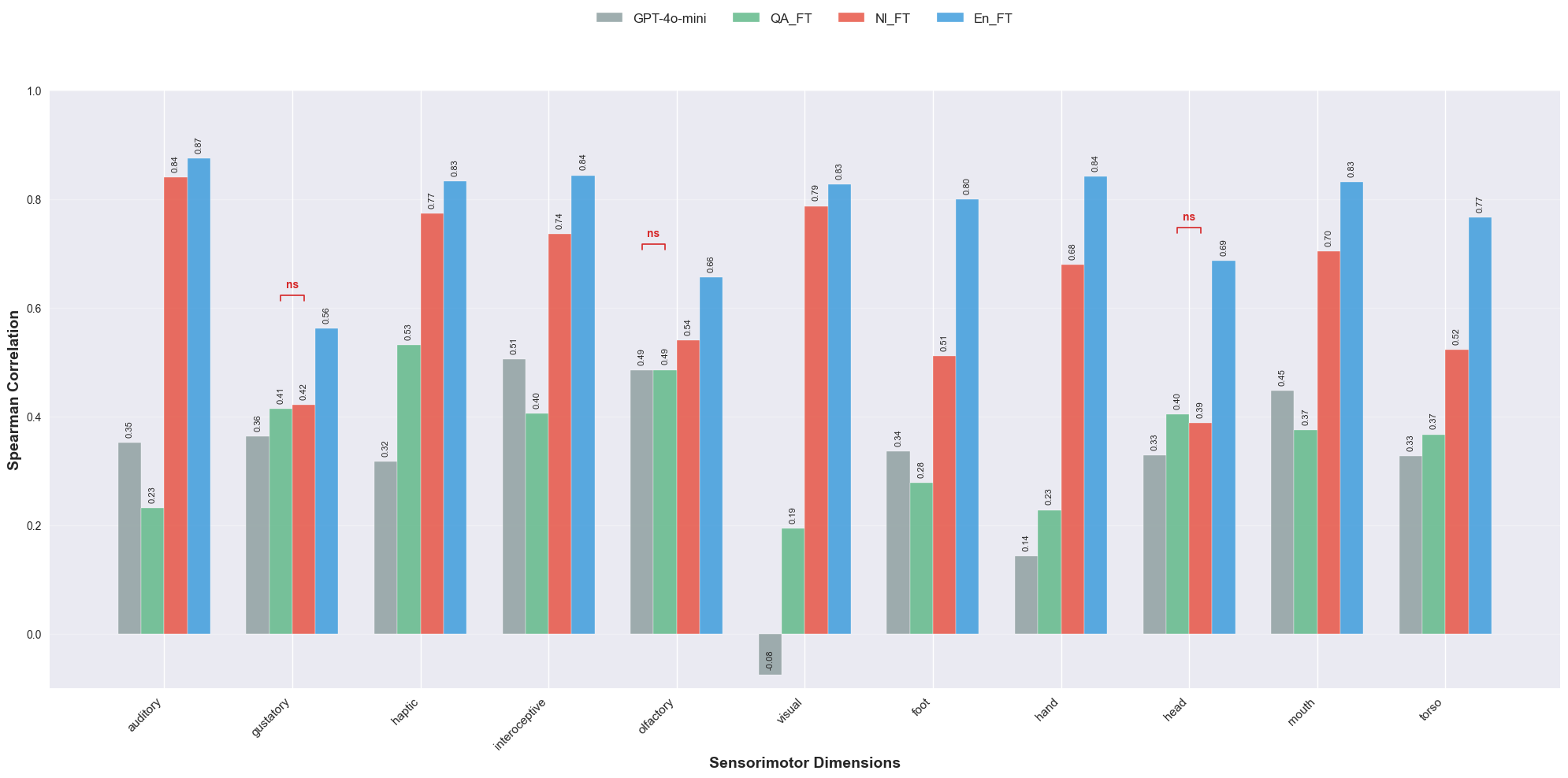}
        \label{fig:en_dim} 
    }
  \vspace{5pt}
    \subfloat[Tested on NL set]{
        \includegraphics[width=0.8\textwidth]{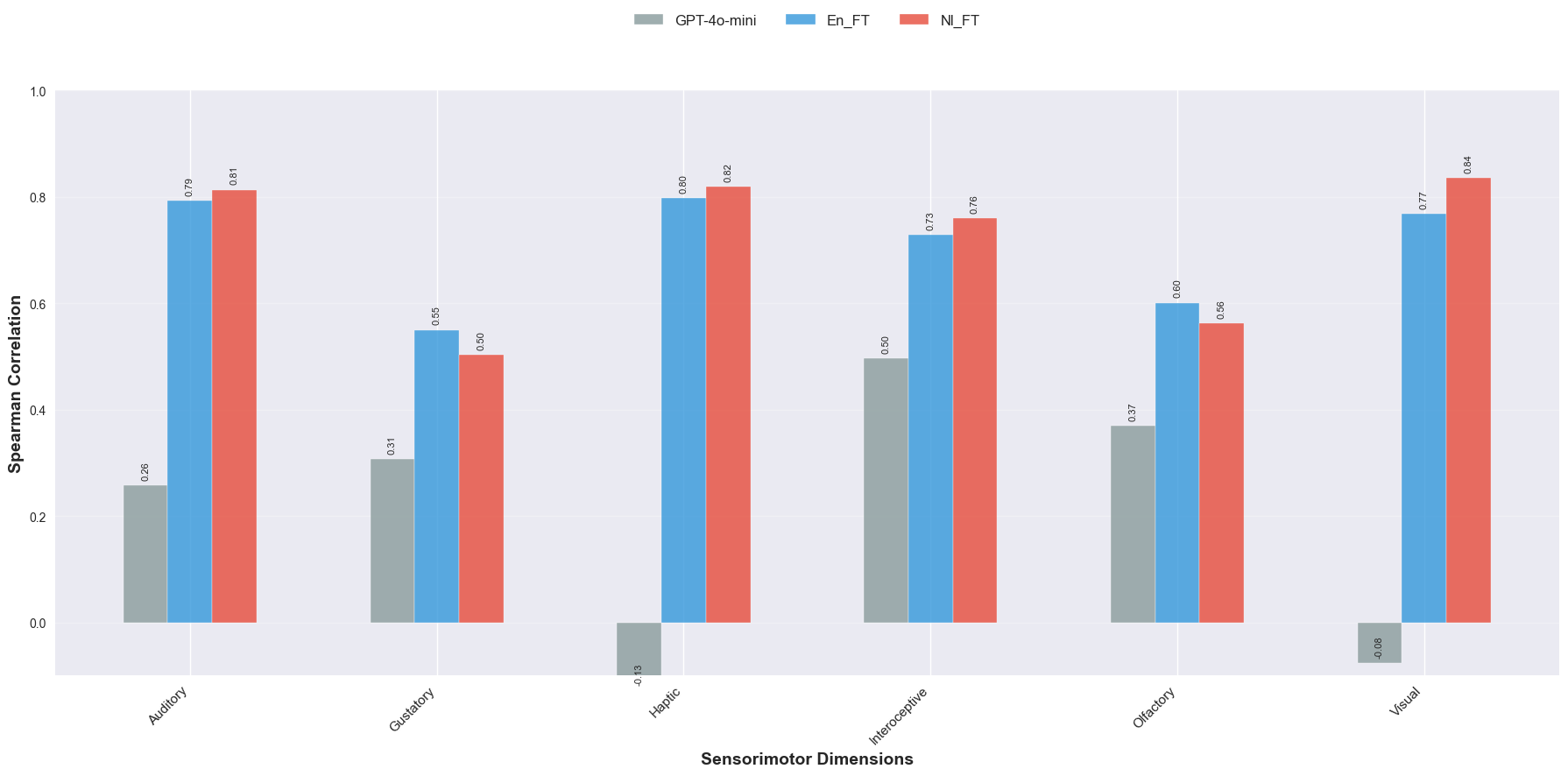}
        \label{fig:dutch_dim}
    }
    \vspace{5pt}
    \subfloat[]{
        \includegraphics[width=0.8\textwidth]{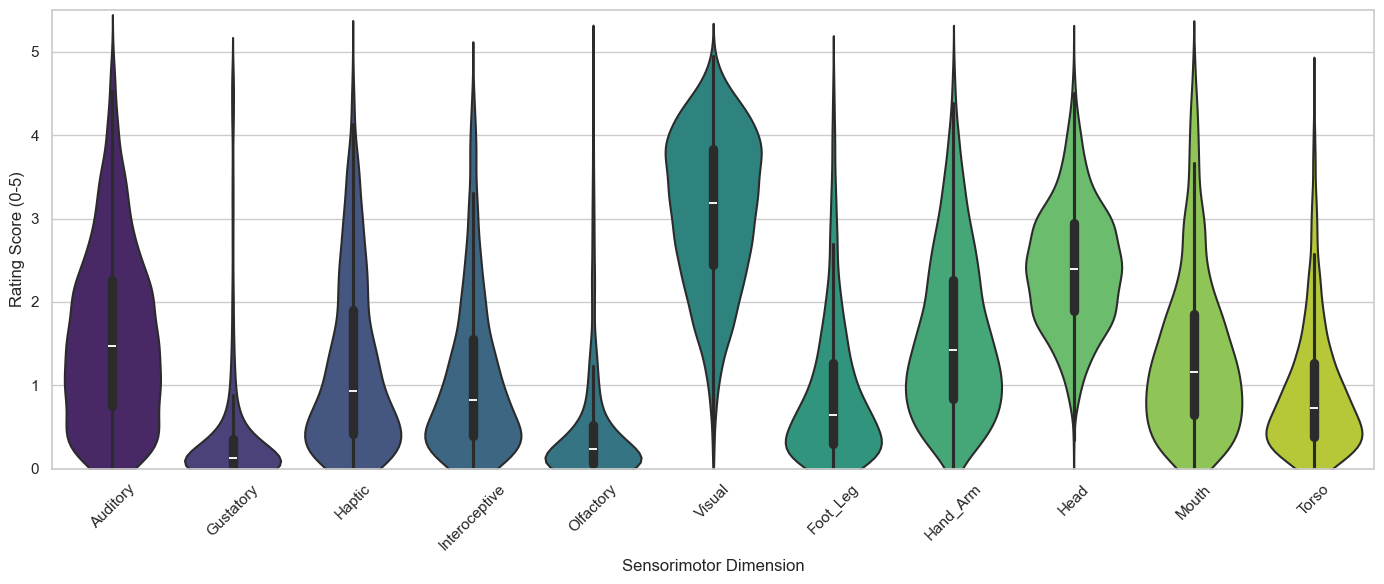}
        \label{fig:dim_human}
    }
    \caption{Dimension-wise representational alignment and the distribution of human ratings. Bar charts display the Spearman’s rank correlation coefficients ($\rho$) between model embeddings and human sensorimotor norms across different dimensions for the (a) English and (b) Dutch evaluation sets. Red brackets labeled ‘ns’ denote non-significant differences ($p \geq 0.5$) from pairwise comparison tests; all other pairwise differences are significant (p < 0.05). (c) illustrates the distribution of human ratings for 2358 concepts across each sensorimotor dimension in the English training set.}
    \label{fig:dimension_results} 
\end{figure}

To dissect the global improvements observed in RSA, we analyzed performance at the level of individual sensorimotor dimensions. As shown in Fig.\ref{fig:en_dim}, the En\_FT and Nl\_FT models exhibited far higher Spearman correlation coefficients than the base model across all 11 dimensions on the English evaluation set. The base model's performance was notably poor, with correlations ranging from a modest 0.51 (Hand) to a markedly low -0.08 (Visual). This pattern confirms the weak sensorimotor grounding of the untuned language model, in line with previous studies \cite{conde_psycholinguistic_2025, xu_large_2025}.

Fine-tuning dramatically rectified this deficit. The En\_FT model achieved correlations above 0.8 in 7 out of 11 dimensions, with its lowest performance (Gustatory, 0.56) still representing a substantial improvement over the base model. Although the Nl\_FT model generally underperformed En\_FT across dimensions, it still significantly surpassed the base model on every dimension. This confirms the general efficacy of fine-tuning and suggests a degree of cross-dimensional generalization, as Nl\_FT improved motor dimension representations (e.g., Hand from 0.14 to 0.68) despite no explicit motor training.

Fig.\ref{fig:dutch_dim} further demonstrates the cross-lingual generalizability of fine-tuning. Mirroring the effective transfer of the Dutch-tuned model (Nl\_FT) to English concepts, the English-tuned model (En\_FT) also achieved strong performance across all dimensions on the Dutch evaluation set. This bidirectional improvement solidifies the conclusion that the sensorimotor structural knowledge acquired through fine-tuning extends across linguistic boundaries. Notably, however, the performance hierarchy shifts with the evaluation language: on the Dutch set, the Nl\_FT model slightly outperformed En\_FT on 4 out of the 6 sensory dimensions. This finding reinforces the conclusion drawn from the RSA analysis that while cross-lingual transfer is robust, optimal dimensional alignment is achieved when the language of the fine-tuning data matches the language of evaluation.

In stark contrast, the QA\_FT model showed inconsistent and limited improvement. On 5 of the 11 dimensions (Auditory, Interoceptive, Olfactory, Foot, Mouth), its performance was not superior to the base model. Even where improvements occurred, they were marginal compared to other fine-tuned models; for example, in the Haptic dimension, QA\_FT reached 0.53 from a base of 0.32, whereas En\_FT and Nl\_FT achieved 0.83 and 0.77, respectively. This indicates that the knowledge gained from a QA-formatted task generalizes poorly to the direct rating prediction task.

Two dimensions warrant specific attention. First, the Visual dimension posed the greatest challenge to the base model, yielding negative correlations in both languages. Second, the Gustatory and Olfactory dimensions showed the most limited gains from fine-tuning (e.g., Gustatory improved from 0.36 to only 0.56 with En\_FT). This limited enhancement can likely be attributed to the constrained variance of human ratings for these experiences. As visualized in the distribution of human ratings (Fig.\ref{fig:dim_human}), scores for Gustatory and Olfactory are heavily concentrated at the low end of the scale (mostly <1), offering the model a less rich signal for learning compared to dimensions like Visual or Hand, which exhibit a much wider rating distribution. This highlights the critical role of training data quality and variance in determining the upper bound of fine-tuning efficacy for specific sensorimotor features.

\subsection{Concept-Level Analysis: Successes and Failure Modes}

Moving beyond structural and dimensional alignment, we analyzed model performance at the finest granularity—individual concepts, using the derived similarity score (from Euclidean distance) as our metric. This analysis reveals that fine-tuning induces not a uniform improvement, but a targeted and systematic reorganization of the model's representational space.

\begin{figure}[htbp]
    \centering
    \subfloat[Tested on EN set]{        
    \includegraphics[width=0.45\textwidth]{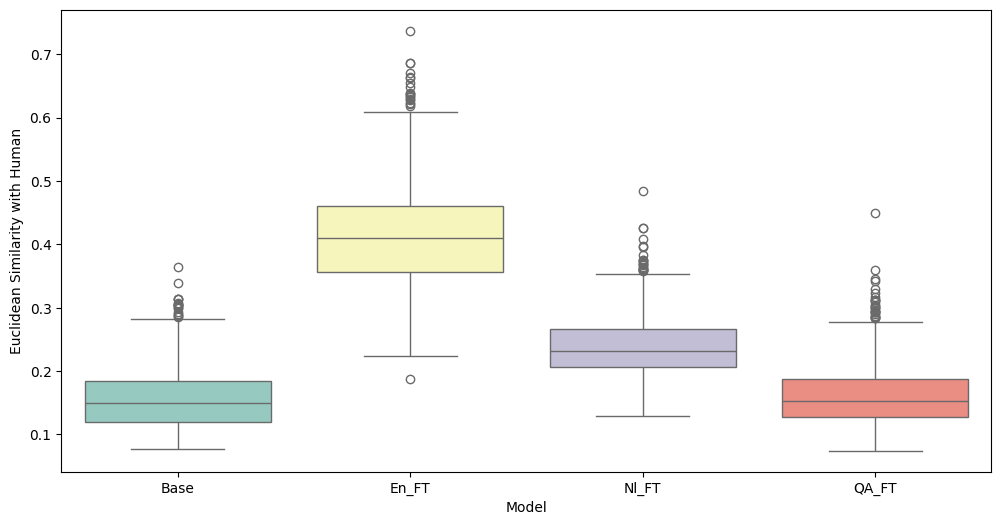}
        \label{fig:en_all} 
    }
    \subfloat[Tested on NL set]{
      \includegraphics[width=0.45\textwidth]{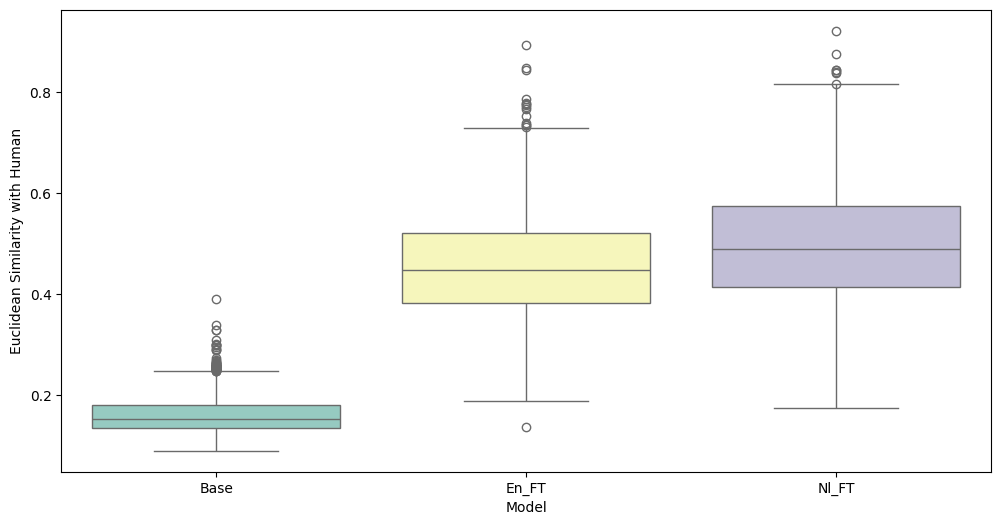}
        \label{fig:dutch_sensory}
    }
   \hfill 
    \subfloat[]{
    \includegraphics[width=0.45\textwidth]{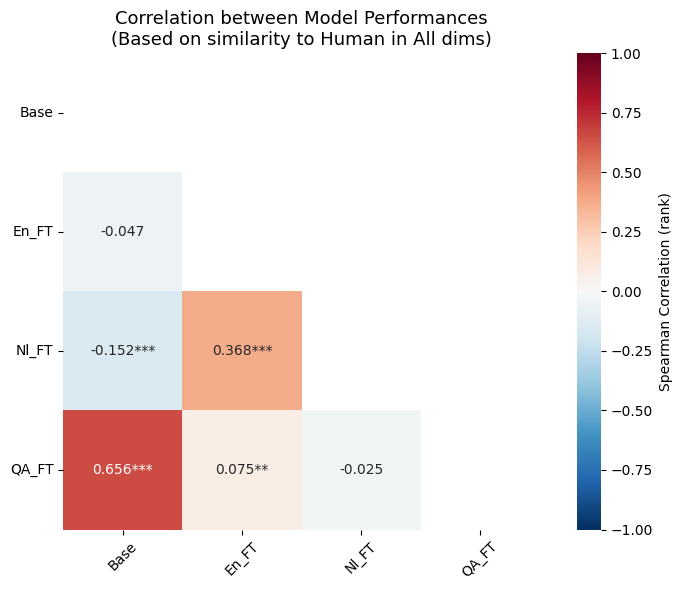}
        \label{fig:corr_model}
    }
    \subfloat[]{
    \includegraphics[width=0.45\textwidth]{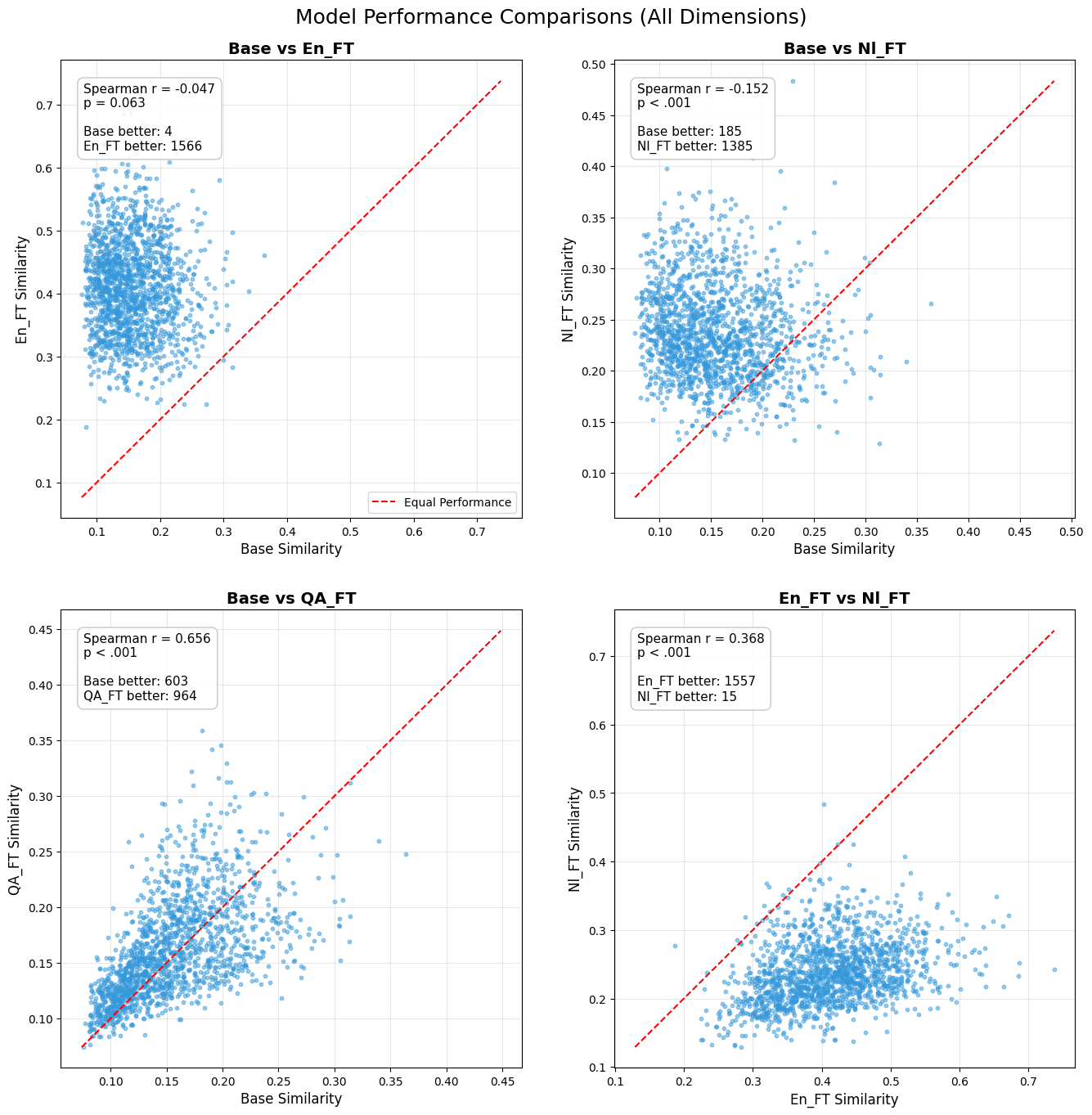}
        \label{fig:scatter}
    }   
     \hfill 
    \subfloat[]{
        \includegraphics[width=0.9\textwidth]{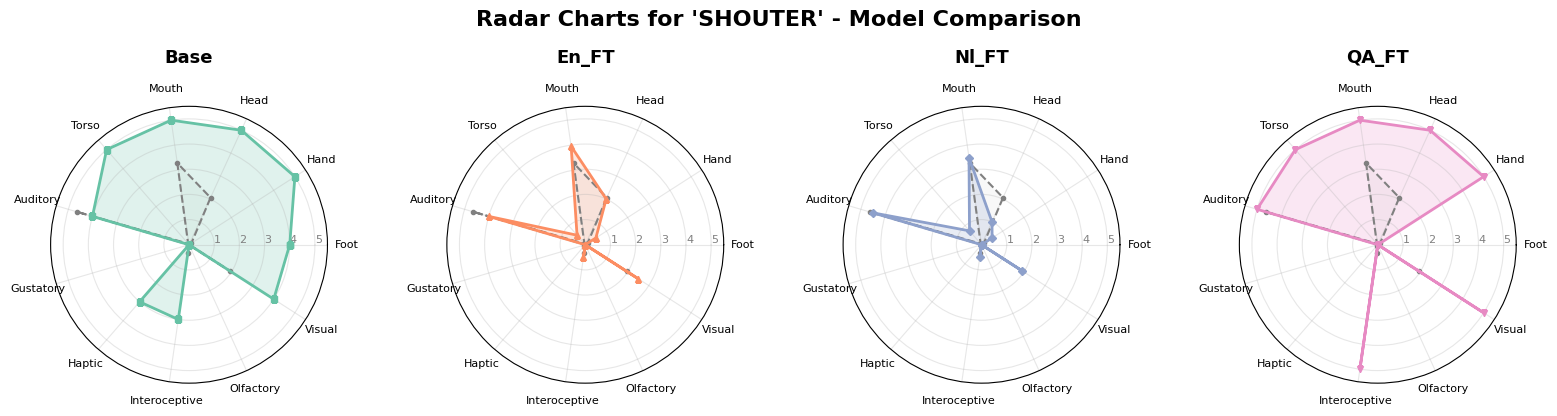}
        \label{fig:shouter_rader}
    }
    \caption{Comparative analysis of word-wise model representations against human ratings. (a, b) Distribution of model-human similarity (calculated via Euclidean distance) for different models on the (a) English and (b) Dutch evaluation sets. (c) Heatmap of inter-model Spearman's $\rho$ correlation, derived from the similarity scores in (a) and (b). (d) Scatter plots contrasting specific model pairs (from left to right, top to down): Base vs. En\_FT, Base vs. Nl\_FT, Base vs. QA\_FT, and En\_FT vs. Nl\_FT. (e) Sensorimotor rating profiles for the exemplar word "SHOUTER". The radar chart compares the predicted rating scores (on a scale of 0–5) for each of the 11 sensorimotor dimensions across the four models (colored lines) with the actual human ratings (gray line).}
    \label{fig:word_results} 
\end{figure}

\paragraph{Hierarchical Performance and Cross-Lingual Transfer}
The distribution of word-level similarity scores (derived from Euclidean distance) reveals a clear hierarchy of model performance (Fig. \ref{fig:en_all} \ref{fig:dutch_sensory}). On the English evaluation set (Fig. \ref{fig:en_all}), the model fine-tuned on English ratings (En\_FT) achieved the highest mean similarity to human ratings (0.41), significantly outperforming the base model (0.16, paired t-test, *p* < .001). This finding directly mirrors its superior performance in both overall structural alignment (RSA) and dimensional correlation, confirming from a word-level perspective that supervised fine-tuning dramatically improves the estimation of precise perceptual strength. 

The model fine-tuned on Dutch ratings (Nl\_FT) also showed significant improvement over the base model on English concepts, again aligning with its effective cross-lingual transfer demonstrated in previous sections. Mirroring this finding on the Dutch evaluation set (Fig. \ref{fig:dutch_sensory}), Nl\_FT performed best, while En\_FT also substantially surpassed the base model. 

In stark contrast, the QA-fine-tuned model (QA\_FT) showed only marginal gains in English (0.16), reiterating its limited efficacy—a conclusion consistent across all levels of our analysis. Thus, the word-level similarity distributions robustly corroborate the established performance hierarchy: matched-language fine-tuning is most effective, followed by cross-lingual fine-tuning, with QA-format fine-tuning yielding minimal gains.

\paragraph{Correlational Evidence for Representational Reorganization and Convergence}
The pattern of inter-model correlations, derived from their word-level similarity scores, provides deeper insight into how fine-tuning reshapes representations (Fig. \ref{fig:corr_model}). Both supervised fine-tuned models (En\_FT and Nl\_FT) showed near-zero to slight negative correlations with the base model (Base vs. En\_FT: $\rho$ = -0.047; Base vs. Nl\_FT: $\rho$ = -0.152). 

This low correlation is critical. It indicates that the relative ranking of concepts from best- to worst-aligned was substantially reshuffled after fine-tuning. A high positive correlation would have suggested a global boost, where all concepts improve uniformly, preserving their original performance order. The observed near-zero correlation, however, signifies a targeted, corrective reorganization. The improvement was not uniform; concepts that were poorly aligned in the base model (the largest errors) may receive the most substantial corrective adjustments from fine-tuning, dramatically improving their ranking. Conversely, concepts that were relatively better aligned in the base model saw less dramatic gains, causing their relative rank to fall. This pattern is visually apparent in the diffuse scatter plot between Base and En\_FT models (Fig. \ref{fig:scatter}), which shows no clear linear relationship, confirming the fundamental recalibration of the representational space.

Furthermore, models fine-tuned with the same task (rating prediction) on different languages (En\_FT vs. Nl\_FT) showed a significant, moderate positive correlation ($\rho$ = .368, p < .001), visualized in their tight linear scatter (Fig. \ref{fig:scatter}). This evidences partial convergence toward a shared representational schema driven by the common learning objective. Conversely, the QA\_FT model’s performance pattern remained strongly correlated with the base model ($\rho$ = .656, p < .001, Fig. \ref{fig:scatter}) but uncorrelated with the supervised models. This confirms that QA-format fine-tuning primarily reinforced the base model's existing representations rather than restructuring them.

\paragraph{Case Study: Sensorimotor Profile of an example Concept} 
The sensorimotor profile of the exemplar word “SHOUTER” illustrates these effects concretely (Fig.\ref{fig:shouter_rader}). The base model produces a relatively flat and inaccurate profile. The En\_FT model corrects this most effectively, showing pronounced, human-like peaks in relevant dimensions such as Auditory and Mouth action. The Nl\_FT model shows a similar but slightly attenuated corrective trend. In particular, the profile of the QA\_FT model remains closest to the base model’s, failing to develop an accurate, peaked representation. This case visually encapsulates the core findings: supervised fine-tuning enables accurate, dimension-specific strength estimation, with cross-lingual convergence, while QA-format tuning does not.

\paragraph{In summary,}
word-level analysis confirms that the primary effect of supervised rating fine-tuning is a targeted, corrective reorganization of the conceptual space. This reorganization leads to superior absolute prediction accuracy, enables positive cross-lingual transfer with partial convergence, and systematically alters performance rankings away from those of the base model. The failure of QA-format fine-tuning to induce this reorganization underscores that the efficacy of parameter adaptation is fundamentally constrained by the alignment between the fine-tuning task and the targeted representational property.

\section{Discussion}
This study systematically investigated whether and how task-specific fine-tuning can enhance the sensorimotor representations in large language models (LLMs). Through a multi-level analysis—encompassing overall structural alignment, dimension-specific accuracy, and word-level precision, we demonstrate that supervised fine-tuning is a highly efficient method for mitigating the well-documented “embodiment gap” in LLMs. Crucially, our findings reveal that the mechanism of improvement is not a simple global amplification but a targeted representational reorganization, with significant implications for our understanding of model plasticity and the development of more grounded AI.

Our results consistently show that fine-tuning with human ratings substantially improves a model's alignment with human sensorimotor experience. At the structural level, fine-tuning significantly increased the similarity between model and human representational spaces (RSA), indicating a global reorganization towards a more human-like semantic structure. At the dimensional level, fine-tuning led to dramatic gains in the accuracy of strength predictions across most sensorimotor features. Notably, dimensions with low variance in human ratings (e.g., Gustatory, Olfactory) showed limited gains, highlighting the dependence of fine-tuning efficacy on the informativeness of the training signal. At the word level, fine-tuning enhanced the absolute numerical accuracy of predictions for individual concepts. The profound insight from this level of analysis is the near-zero correlation between the performance rankings of fine-tuned and base models, which decisively shows that improvement comes from a corrective recalibration of the most misaligned concepts, not from a uniform boost.

Furthermore, our experiments on generalizability revealed a clear hierarchy: fine-tuning generalizes robustly across languages (with a performance advantage for language-matched training), but poorly across task formats. Models fine-tuned on Dutch ratings improved English predictions, and vice versa. However, a model fine-tuned on a QA task failed to transfer effectively to rating prediction, performing similarly to the base model. This underscores that the learning objective—not just exposure to human judgments—is critical for shaping the desired representational change. Finally, we observed cross-dimensional generalization: models fine-tuned only on sensory data also improved their representations of motor features, suggesting an interconnectedness in the learned sensorimotor semantic space.

A key theoretical insight of this work is that fine-tuning drives a process of targeted representational reorganization. The near-zero Spearman correlation between pre- and post-fine-tuning performance rankings rules out a simple "global boost." Instead, gradient updates selectively correct the largest errors: concepts where prior textual predictions deviated most from human experience receive the strongest learning signal, leading to dramatic improvement and a reshuffled performance hierarchy. This mechanism aligns with theories of human perceptual learning, where feedback targets areas of greatest discrepancy. In contrast, the QA\_FT model's strong correlation with the base model shows that without a task objective aligned with the target representation, learning merely refines—rather than reorganizes—the existing flawed structure.

Our findings offer a constructive pathway to address the embodiment gap identified by \cite{conde_psycholinguistic_2025} and \cite{xu_large_2025}. While they concluded that LLMs lack grounded sensorimotor representations due to their text-only training, we show that this gap can be substantially narrowed through efficient parameter adaptation using limited human ratings, without the need for costly multimodal pre-training. This demonstrates a remarkable plasticity in LLMs; their representations are not fixed but can be steered towards more embodied patterns through targeted supervision. The partial cross-lingual convergence we observed suggests that this plasticity operates on abstract, amodal structural patterns that can be accessed via different linguistic instantiations.

Several limitations point to future research. First, the limited improvement for dimensions like Gustatory and Olfactory is likely attributable to data limitations, particularly the low variance and constrained range of human ratings in these domains, which suggests the need for more nuanced or extensive human ratings that better capture the full spectrum of these sensory experiences. Second, we explored a limited set of model architectures and fine-tuning scales; investigating the dose-response relationship and scaling laws of sensorimotor fine-tuning is a logical next step.

To fundamentally address the issue of task-specific representations and cultivate more generalized embodied knowledge, future work could incorporate neuroscientific data such as fMRI or EEG recordings of human brain activity during sensorimotor processing. By using these neural signals as additional supervision signals during training—for instance, by aligning the model's internal activations with patterns of brain activity evoked by the same concepts, we may guide the model to develop representations that more closely resemble the integrative and multimodal nature of human embodied cognition. This neuro-aligned approach has the potential to move beyond the limitations of task-specific fine-tuning, forcing the model to learn the underlying neural principles of sensorimotor integration rather than merely optimizing for superficial task performance.

\section{Methods}
\subsection{Inclusion and ethics}  
The study involves the collection of data from LLMs and the use of secondary human-participant data \cite{lynott_lancaster_2020,speed_dutch_2022,yang_does_2025}. For the human-participant data, ref. 1 noted that the study followed the ethical guidelines and protocols established by the British Psychological Society. Ethical approval for the study reported in ref. 2 was granted by the Lancaster University Research Ethics Committee.

\subsection{Psycholinguistic Norms}
We employed three distinct sets of human behavioral data to fine-tune and evaluate models across sensorimotor, cross-lingual, and cross-task settings.

\paragraph{English Sensorimotor Norms.}
We used the Lancaster Sensorimotor Norms \cite{lynott_lancaster_2020} as our primary dataset of human perceptual strength ratings. These norms provide ratings on a scale from 0 (no experience) to 5 (strong experience) for 39,710 English words across 11 dimensions: six sensory (Visual, Haptic, Auditory, Olfactory, Gustatory, Interoceptive) and five motor (Foot/Leg, Hand/Arm, Mouth, Torso, Head). These norms have been validated as a reliable representation of conceptual sensorimotor information \cite{wingfield_sensorimotor_2023} and are widely used in related research \cite{conde_psycholinguistic_2025, xu_large_2025}.

\paragraph{Dutch Sensory Norms and Bilingual Dataset Construction.}
To enable cross-lingual evaluation, we utilized the Dutch sensory modality norms \cite{speed_dutch_2022}, which provide perceptual strength ratings for over 24,000 Dutch words across six sensory modalities (Auditory, Gustatory, Haptic, Olfactory, Visual, Interoceptive). 

To construct a high-quality, aligned bilingual dataset, we first translated the Dutch words into English and performed back-translation using the DeepL API, retaining only word pairs that matched exactly in the original and back-translated Dutch forms and were non-identical cognates. This resulted in 10,677 reliable translation pairs. We then identified the 7,861 pairs that also existed in the Lancaster norms, ensuring full sensory modality alignment. From this pool, 3,930 word pairs (50\%) were randomly selected for this study and split into a training set (2,358 words, 60\%) and a held-out test set (1,572 words, 40\%).

\paragraph{Question-Answering (QA) Task Data.}
To fine-tune and evaluate models on a different task format, we employed the PerceptualQA dataset \cite{yang_does_2025}. This benchmark comprises 1,400 multiple-choice questions designed to probe embodied knowledge, covering subtasks across five visual domains (200 questions each) and four other sensory modalities (Auditory, Tactile, Gustatory, Olfactory; 100 questions each). The dataset was split into a training set (800 samples, 60\%) and a test set (600 samples, 40\%), preserving the original label distribution.

\subsection{Prompt Design}
Prompts were designed as self-contained, atomic tasks to convert the human rating data into a format suitable for model training and evaluation, ensuring consistency across all stages.

\paragraph{Rating Prediction Prompt (English \& Dutch).}
For the sensorimotor rating task, we used a dedicated template per dimension. Each prompt constituted an independent example. For instance, to query the \textit{Visual} strength of the word ``apple'', the prompt was formatted as:

\begin{quote}
    \textit{``Complete the following task as a native speaker of English. Please indicate how strongly the concept is experienced by seeing on a scale from 0 (not at all) to 5 (greatly), with the midpoint representing moderate strength. The concept is: `apple'. Only answer with a number from 0 to 5.''}
\end{quote}

The core experiential description (\textit{seeing} in the example) was customized for each of the 11 Lancaster dimensions (e.g., ``hearing'' for Auditory, ``performing an action with the hand/arm'' for Hand\_Arm). For Dutch prompts, the phrase ``as a native speaker of English'' was replaced with ``as a native speaker of Dutch''. This identical, atomic prompt structure was used to create every individual training and testing example. For model fine-tuning, the complete dataset was constructed by applying this template to every combination of a training word and its relevant dimension(s) (e.g., 2,358 words $\times$ 11 dimensions for English).

\paragraph{Question-Answering (QA) Prompt.}
For the QA task, a standardized multiple-choice format was employed, derived from the PerceptualQA dataset:

\begin{quote}
    \textit{``Read the question and the answer choices below. Choose the single best answer and reply with exactly one uppercase letter from the set \{A, B, C, D\} --- and nothing else.\\
    Question \{index\}: \{question\}\\
    Choices: \{options\}''}
\end{quote}

The placeholders \texttt{\{index\}}, \texttt{\{question\}}, and \texttt{\{options\}} were populated directly from the dataset to form each training or test example.

\subsection{Models and Fine-tuning Procedure}

\paragraph{Base Model.}
We used \texttt{gpt-4o-mini-2024-07-18} accessed via the official OpenAI API as our foundation. This model was selected due to the strong performance of the GPT-4 family and to enable direct comparison with prior work \cite{conde_psycholinguistic_2025, xu_large_2025}.

\paragraph{Fine-tuned Models.}
Using the base model, we created three primary fine-tuned variants via OpenAI’s dedicated fine-tuning service:
\begin{itemize}
    \item \textbf{En\_FT:} Fine-tuned on the complete set of English rating-prediction examples. This comprised prompts for all 11 dimensions across each of the 2,358 training words (totaling $2,358 \times 11 = 25,938$ examples).
    \item \textbf{Nl\_FT:} Fine-tuned on the Dutch sensory rating-prediction examples. This comprised prompts for all 6 sensory dimensions across each of the 2,358 Dutch training words (totaling $2,358 \times 6 = 14,148$ examples).
    \item \textbf{QA\_FT:} Fine-tuned on the English PerceptualQA multiple-choice examples using the QA prompt format.
\end{itemize}

During evaluation, the same per-word, per-dimension prompting strategy was employed. For a given model and test set, each word was queried independently across the relevant dimensions to generate a complete rating vector. To minimize stochasticity and favor deterministic and comparable outputs, the temperature parameter was set to 0 for all model calls during evaluation.

\subsection{Analysis}

We employed a multi-level analysis framework to comprehensively evaluate the alignment between model predictions and human sensorimotor ratings, focusing on overall representational structure, dimension-specific accuracy, and concept-level precision.

\paragraph{Overall Representational Similarity Analysis (RSA).}
To assess the global structural alignment between model-derived and human-derived semantic spaces, we performed Representational Similarity Analysis (RSA). For each model and for the human ratings, we constructed a Representational Dissimilarity Matrix (RDM) based on the complete rating vectors for each concept. Cosine distance was used to compute dissimilarity between concept vectors after standardizing the feature vectors. The model's performance at this level was quantified as the Spearman rank correlation coefficient ($\rho$) between the upper-triangular portions of the model RDM and the human RDM. To statistically compare the RSA performance between two models (e.g., a fine-tuned model vs. the base model), we employed a bootstrap resampling test. In each of 200 iterations, concepts were resampled with replacement, and the difference in correlation with the human RDM ($\Delta \rho$) was computed for the model pair. Statistical significance was established if the 95\% confidence interval of the bootstrapped differences excluded zero.

\paragraph{Dimension-Level Correlation Analysis.}
To quantify alignment for specific sensorimotor features, we calculated the Spearman's rank correlation ($\rho$) between model-predicted and human ratings for each of the 11 dimensions separately. To test for significant performance differences between models on a given dimension (e.g., En\_FT vs. Base on the Visual dimension), we used Steiger's Z-test for dependent correlations with a common variable (the human ratings), implemented via the \texttt{cocor} R package \cite{steiger_tests_1980}.

\paragraph{Word-Level Accuracy Analysis.}
To evaluate the absolute numerical accuracy of predictions for individual concepts, we computed the Euclidean distance between the model-generated 11-dimensional rating vector and the corresponding human rating vector for each word. To facilitate interpretation and visualization, this distance ($d$) was converted to a similarity score bounded between 0 and 1: $similarity = 1 / (1 + d)$. Distributions of these similarity scores were analyzed for each model across three subsets: “All”, “Motor”, and “Sensory” concepts. To statistically compare the performance of two models at the word level (e.g., fine-tuned vs. base), we treated the per-concept similarity scores (or their underlying Euclidean distances) as paired samples and applied paired-sample t-tests.

\bibliographystyle{unsrt}  
\bibliography{references} 

\end{document}